\documentclass[letterpaper,10pt,conference]{ieeeconf}
\IEEEoverridecommandlockouts                              

\usepackage[T1]{fontenc}
\usepackage[utf8]{inputenc}
\usepackage{siunitx}
\usepackage{algorithm}
\usepackage{algpseudocode}
\usepackage{algpseudocodex}
\usepackage{array}
\usepackage{amsmath}
\usepackage{amssymb}
\usepackage{booktabs}
\usepackage{color}
\usepackage{dsfont}
\usepackage{epsfig}
\usepackage{graphics}
\usepackage[utf8]{inputenc}
\usepackage{listings}
\usepackage{makecell}
\usepackage{multirow}

\usepackage{pifont}
\usepackage{times}
\usepackage{xcolor}
\usepackage{hyperref}
\usepackage{cleveref}
\usepackage{mathtools}
\usepackage{array} 
\usepackage{times}
\usepackage{epsfig}
\usepackage{algorithm}
\usepackage{tabularx}
\usepackage{makecell}
\usepackage{graphicx}
\usepackage{amsmath}
\usepackage{amssymb}
\usepackage{booktabs}
\usepackage{tabularx}
\usepackage{booktabs}
\usepackage{makecell} 
\usepackage{array}    
\usepackage{multirow} 
\usepackage{xcolor}
\usepackage{colortbl} 

\usepackage{pifont}

\usepackage{array}
\usepackage{amsmath}
\usepackage{amssymb}
\usepackage{booktabs}
\usepackage{color}
\usepackage{dsfont}
\usepackage{epsfig}
\usepackage{graphics}
\usepackage[utf8]{inputenc}
\usepackage{listings}
\usepackage{makecell}
\usepackage{multirow}
\usepackage{pifont}
\usepackage{times}
\usepackage{xcolor}
\usepackage{hyperref}
\usepackage{cleveref}
\usepackage{times}
\usepackage{latexsym}

\title{\LARGE \textbf{ViLaD: A Large Vision Language Diffusion Framework for End-to-End Autonomous Driving}}

\author {
    Can~Cui, 
    Yupeng~Zhou,
    Juntong~Peng,
    Sung-Yeon~Park,
    Zichong~Yang,
    Prashanth~Sankaranarayanan, \\
    Jiaru~Zhang,
    Ruqi~Zhang,
    and~Ziran~Wang
\thanks{C. Cui, Y. Zhou, J. Peng, S.-Y. Park, Z. Yang, P. Sankaranarayanan, J. Zhang, R. Zhang, and Z. Wang are with the College of Engineering, Purdue University, West Lafayette, IN 47907, USA.}
\thanks{R. Zhang is with the Computer Science Department, Purdue University, West Lafayette, IN 47907, USA.} 
\thanks{Corresponding author: J. Zhang, email: {\tt\small jiaru@purdue.edu.}}%
}
\begin{document}






\markboth{Journal of \LaTeX\ Class Files,~Vol.~14, No.~8, August~2015}%
{Shell \MakeLowercase{\textit{et al.}}: Bare Demo of IEEEtran.cls for IEEE Journals}

\maketitle%

\begin{abstract}
End-to-end autonomous driving systems built on Vision Language Models (VLMs) have shown significant promise, yet their reliance on autoregressive architectures introduces some limitations for real-world applications. The sequential, token-by-token generation process of these models results in high inference latency and cannot perform bidirectional reasoning, making them unsuitable for dynamic, safety-critical environments. To overcome these challenges, we introduce ViLaD, a novel Large Vision Language Diffusion (LVLD) framework for end-to-end autonomous driving that represents a paradigm shift. ViLaD leverages a masked diffusion model that enables parallel generation of entire driving decision sequences, significantly reducing computational latency. Moreover, its architecture supports bidirectional reasoning, allowing the model to consider both past and future simultaneously, and supports progressive easy-first generation to iteratively improve decision quality. We conduct comprehensive experiments on the nuScenes dataset, where ViLaD outperforms state-of-the-art autoregressive VLM baselines in both planning accuracy and inference speed, while achieving a near-zero failure rate. Furthermore, we demonstrate the framework's practical viability through a real-world deployment on an autonomous vehicle for an interactive parking task, confirming its effectiveness and soundness for practical applications. 
\end{abstract}


\section{Introduction}
Autonomous driving represents one of the most challenging applications of artificial intelligence, requiring real-time perception, reasoning, and decision-making in dynamic, safety-critical environments~\cite{cui2023survey}. 
As an alternative to traditional module-based pipelines, end-to-end autonomous driving aims to use a single model to directly map raw sensor data to driving commands, which has become a popular autonomous driving paradigm.
Recent advances have shown remarkable promise for end-to-end autonomous driving by enabling vehicles to process multimodal sensory inputs and generate human-interpretable driving decisions through Vision Language Models (VLMs)~\cite{tian_drivevlm_2024,xu_drivegpt4_2023,cui_drive_2024}. 
These approaches leverage the powerful reasoning capabilities of VLMs to bridge the gap between raw sensor data and high-level driving commands, offering better interpretability and flexibility in autonomous driving systems.

However, current VLM-based end-to-end autonomous driving systems significantly rely on autoregressive architectures that generate decisions through sequential next-token prediction. 
This autoregressive paradigm, while successful for many natural language processing tasks, presents some fundamental limitations when applied to the time-sensitive and spatially complex domain of real-world driving scenarios. 
These models generate decisions sequentially, token-by-token, which is \textit{computationally inefficient} and \textit{conceptually inflexible}: 
First, the sequential generation process creates significant inference latency, as each token must be generated one by one, making real-time decision-making challenging for safety-critical scenarios; 
Second, the left-to-right generation paradigm limits the model's ability to consider global spatial relationships and future trajectory implications simultaneously, potentially leading to suboptimal driving decisions that lack comprehensive scene understanding. 
In the dynamic and unpredictable environment of the road, this lack of foresight and adaptability may cause a significant challenge to safety and reliability;
Moreover, autoregressive VLMs suffer from the ``reversal curse'' phenomenon~\cite{berglund2024reversalcursellmstrained}, where models trained on sequential patterns struggle with tasks requiring bidirectional reasoning or reverse inference. In autonomous driving, this limitation causes difficulty in reasoning backward from desired outcomes (e.g., ``To avoid the congested traffic flow, I should exit the highway at the next exit'') or integrating future planning constraints into current decisions (e.g., ``Stop at the next intersection'').


To address these fundamental limitations, we propose a fundamental paradigm shift for end-to-end autonomous driving from autoregressive VLMs to the Large Vision Language Diffusion (LVLD) models \cite{you2025lladavlargelanguagediffusion}. 
These kinds of models rely on a masked diffusion generation paradigm. Specifically, the text is generated by starting with a fully masked sequence, which is then iteratively filled in with tokens predicted by a model trained to reverse this masking process. 
Therefore, it offers several advantages in end-to-end autonomous driving:
\begin{itemize}
    \item \textbf{Efficient Parallel Generation}. Unlike sequential token prediction, diffusion models generate the decision sequences simultaneously, significantly reducing inference latency.
    \item  \textbf{Bidirectional Reasoning}. The iterative bidirectional generation process enables models to consider both past context and future influences comprehensively.
    \item \textbf{Easy-First Generation Patterns}. The model can adaptively focus on easier aspects of the driving task first, then progressively address more complex aspects.
\end{itemize}

With the utilization of LVLD models, we introduce \textbf{ViLaD} (\textbf{Vi}sion \textbf{La}nguage \textbf{D}iffusion), a novel end-to-end autonomous driving framework that leverages the masked diffusion generation method. This technique trains a neural network model to reconstruct a complete sequence of driving actions from a version where parts of the output are deliberately ``masked.'' During inference, the model begins with a fully masked (excepted fixed tokens) sequence of decision tokens and iteratively ``unmasks'' the masked tokens in parallel over several steps to produce the final output. It demonstrates a successful end-to-end implementation of the LVLD autonomous driving system from the benchmark comprehensive experiment to a real-world vehicle deployment, which is the \textbf{first of its kind} to the best of our knowledge. 
The main contributions of this paper are summarized as follows:

\begin{itemize}
    \item We are the first to successfully adapt LVLD models for end-to-end autonomous driving tasks, establishing a new paradigm that moves beyond autoregressive generation to enable more efficient and comprehensive decision-making in safety-critical driving scenarios.
    \item We develop and analyze novel inference optimization strategies tailored for diffusion-based driving models, including a Fixed Pattern (FP) training and inference method and a confidence-based parallel decoding approach. These techniques significantly improve inference speed while maintaining high accuracy by optimizing the generation process.
    \item We perform comprehensive experiments on the nuScenes dataset, demonstrating that our ViLaD framework outperforms existing VLM-based methods in autonomous driving, achieving a state-of-the-art average L2 error of 1.81 m and a near-zero failure rate.
    \item We successfully deploy our framework on a real-world vehicle, bridging the gap between research and practical application. In a real-time, on-board interactive parking task, our model demonstrated its practical value and readiness by achieving a 95.67\% success rate with an average inference latency of only 0.96 seconds.
\end{itemize}




\section{Related Works}
\subsection{VLM-based End-to-end Autonomous Driving}

VLMs are being increasingly applied to end-to-end autonomous driving systems. An effective application is using VLMs to autoregressively predict future waypoints or actions. For example, DriveGPT4 was proposed as a GPT-style model to predict encoded future scene tokens~\cite{huang2024drivegpt}, and another work has focused on generating multi-step control actions guided by trajectory predictions~\cite{wu2022trajectory}.

Beyond simple generation, the reasoning capabilities of VLMs are being leveraged to build more robust and interpretable autonomous driving systems. To this end, researchers have developed question-answering frameworks where the model can explain its decisions~\cite{xu_drivegpt4_2024}. The reasoning process has been further improved through Chain-of-Thought (CoT) prompting, where the model generates intermediate logical steps before reaching a final decision, a technique explored by DriveCoT~\cite{wang_drivecot_2024}. VLM frameworks like EMMA and OpenEMMA have also demonstrated effectiveness and generalizability across challenging driving scenarios~\cite{hwang2024emma, xing2025openemma}. Furthermore, some approaches use VLMs in a dual-process framework, combining a powerful but slower VLM for analytical reasoning with a faster, lightweight model for real-time control, mimicking human cognition to enhance performance and adaptability~\cite{mei_continuously_2024}. Other works use VLMs to provide high-level guidance or ``meta-actions'' to a separate end-to-end driving model, enhancing its scene understanding and ability to handle novel situations~\cite{jiang2024senna, guo2025vdt, fu2025orion}. However, these current VLM-based autonomous driving systems heavily rely on autoregressive architectures that generate decisions through sequential, next-token prediction~\cite{chen2024next}. This paradigm introduces inherent limitations; the token-by-token generation process leads to high inference latency, and its unidirectional nature restricts the model's ability to consider future constraints, making it less suitable for dynamic, safety-critical environments~\cite{berglund2024reversalcursellmstrained}.

\subsection{The Development of Diffusion Language Model}

Diffusion models, which have demonstrated remarkable success in visual domains by generating high-quality images~\cite{sohl2015deep,ho2020denoising,song2020score}, have recently become popular, though not yet fully established, alternative for LLMs~\cite{you2025lladavlargelanguagediffusion,nie2025large}. The application of these models to natural language processing has required solving the challenge of adapting a framework designed for continuous data, like images, to the discrete nature of text~\cite{nie2025large}. Current prominent approaches include replacing the continuous diffusion process with discrete counterparts that define new forward and reverse dynamics for textual data~\cite{austin2021structured}.

This has led to the development of various discrete diffusion models, among which masked diffusion models (MDMs) have shown particular promise and scalability~\cite{nie2025large,lou2023discrete}. MDMs operate through a process of progressively masking tokens in a sequence and then training a model to predict these masked tokens, thereby reversing the process~\cite{nie2025large}. Recent research has demonstrated the potential of scaling MDMs to sizes comparable to prominent autoregressive LLMs~\cite{nie2025large,nie2025scalingmaskeddiffusionmodels}. For instance, LLaDA, an 8-billion parameter masked diffusion model, has achieved performance on par with leading LLMs such as LLaMA 3 on various benchmarks~\cite{nie2025large}. This work has shown that MDMs can be effectively scaled and can exhibit strong in-context learning and instruction-following capabilities. Furthermore, the bidirectional nature of MDMs allows them to address inherent limitations of autoregressive models, such as the ``reversal curse,'' where models struggle with tasks that require reasoning in a non-sequential order~\cite{berglund2024reversalcursellmstrained,nie2025large}. Additionally, some approaches have focused on fine-tuning existing autoregressive models using an MDM framework~\cite{gong2024scaling}. While still an emerging area, the progress in scaling and refining diffusion-based language models suggests they are a viable and promising direction for the future of generative AI.

\begin{figure*}[!t]
    \centering
    \includegraphics[width=0.9\textwidth]{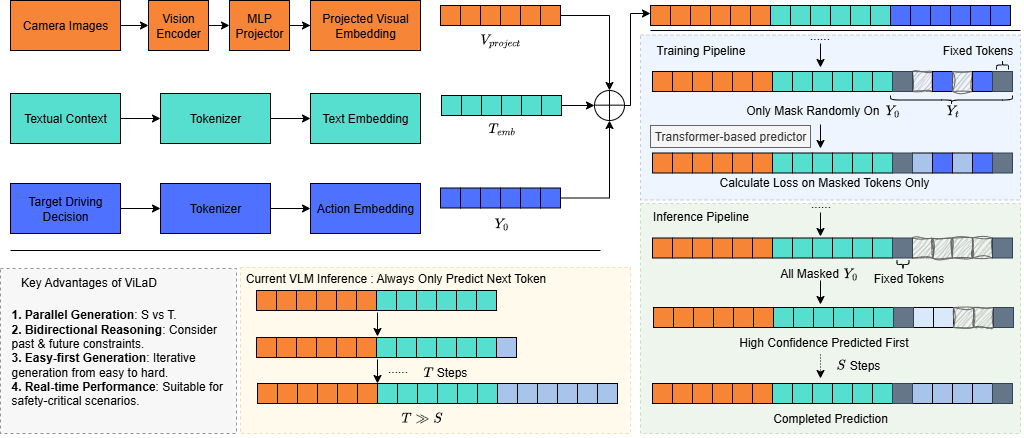}
    \caption{This diagram shows the complete ViLaD framework for end-to-end autonomous driving: The left side illustrates multimodal input processing, where camera images and textual context are encoded and concatenated with action embeddings. The right side demonstrates the core innovation with training and inference pipelines: during training, only driving decision tokens are masked while visual/textual inputs remain unchanged; during inference, the model progressively demasks predictions starting with high-confidence decisions first. The bottom highlights ViLaD's key advantages over autoregressive VLMs, including parallel generation, bidirectional reasoning, easy-first generation, and computational efficiency.}
    \vspace{-6mm}
    \label{fig:main}
\end{figure*}

\section{ViLaD: Vision Language Diffusion}

We present ViLaD, a novel architecture that leverages masked diffusion models for end-to-end autonomous driving. Our architecture consists of four key components: a vision encoder for camera-based perception, a multilayer perceptron (MLP) connector for multimodal alignment, and a diffusion-based language backbone for decision generation.

\subsection{Motivation}
\subsubsection{Limitations of Autoregressive Vision Language Models.}
Current state-of-the-art VLMs for autonomous driving mostly employ autoregressive generation, where the probability of generating a driving action sequence $y = (y_1,y_2, ... , y_{T})$ with length $T$ given visual and text input $x$ is factorized as:
\begin{equation}
    P(y|x) = \prod_{i=1}^TP(y_i|y_1,y_2, ... , y_{i-1},x).
\end{equation}

This formulation introduces several fundamental limitations for autonomous driving. Specifically, the sequential dependency requires $T$ forward passes for generating a complete driving decision.
Therefore, when the length of the driving sequence is large, the inference will require a large time cost, and the inference latency will be unacceptable. For real-time autonomous driving systems requiring seconds to respond~\cite {cui2024onboardvisionlanguagemodelspersonalized}, this sequential bottleneck becomes prohibitive.

In addition, the next-token prediction from the autoregressive model inherently constrains the model to left-to-right information flow, preventing the integration of future trajectory constraints into current decision-making. Mathematically, this limitation can be expressed as:

\begin{equation}
    P(y_i|y_1,y_2, ... , y_{i-1},x) \neq P(y_i|y_1,y_2, ... , y_{i-1},y_{i+t},x).
\end{equation}



\subsubsection{Vision Language Diffusion Models in Autonomous Driving.}
Diffusion models address these fundamental limitations through parallel generation and bidirectional reasoning. The reverse process generates all tokens simultaneously according to:
\begin{align}
p_{\theta}(y_0 | y_t, x) = \prod_{i\in\{i| y_t^i\ = [M]\}} p_{\theta}(y_0^i | y_t, x).
\end{align}
This reduces the number of inference steps from $T$ to $S$,
where $S$ represents the number of diffusion steps. Since $S \ll T$ typically, this provides a significant speedup, enabling real-time performance for autonomous driving applications. 

Apart from that, the diffusion formulation enables bidirectional dependencies through:
\begin{equation}
    p_{\theta}(y_0^i | y_t, x) = f_{\theta}(y_t, x),
\end{equation}
where $f_\theta$ can access the complete generated sequence context without causal constraints. This mathematical property allows the model to consider both historical observations and future planning constraints.



A particularly compelling advantage demonstrated by our ViLaD is the progressive, easy-to-hard generation pattern enabled by the iterative denoising process. The confidence-guided remasking strategy allows the model to focus on high-confidence predictions first, then progressively generate more challenging aspects. Mathematically, this can be expressed as the model learning a difficulty-aware generation order $\pi(i)$ where easier tokens are predicted with higher confidence scores $p_\theta(y^i_{o}|y_t, x)$ at earlier denoising steps. For example, when approaching a stop sign, the model can first establishe the certain decision ``vehicle will stop at stop sign location $(x_{stop}, y_{stop})$'' with high confidence,  then generates the approach waypoints ``vehicle follows path $(x_1, y_1) \rightarrow (x_2, y_2) \rightarrow (x_{stop}, y_{stop})$'' representing the trajectory leading to the stop sign, and finally generates the most challenging decisions about post-stop navigation such as ``turn left and merge into target lane'' with detailed waypoint sequences. Mathematically, this can be formalized as a confidence-ordered generation sequence where $p_\theta(y_{stop}|y_{ts}, x)$ at early denoising steps, followed by $p_\theta(y_{approach\_path}|y_{ta}, x)$ for pre-stop trajectory planning, and finally $p_\theta(y_{post\_action}|y_{tp}, x)$ for complex post-stop maneuvers, where $p_\theta(y_{stop}|y_{ts}, x) \geq p_\theta(y_{approach\_path}|y_{ta}, x) \geq p_\theta(y_{post\_action}|y_{tp}, x)$.

\subsection{Overall Architecture Design}
ViLaD follows an encoder-decoder architecture where visual inputs are first processed through a vision encoder, then projected into the language embedding space through an MLP connector, concatenated with textual driving context, and finally processed by a transformer-based mask predictor to predict masks. Unlike traditional autoregressive VLMs that generate tokens sequentially, ViLaD generates complete driving action sequences through parallel diffusion-based generation, enabling real-time performance while maintaining bidirectional reasoning capabilities.

The architecture takes as input multimodal driving data: camera images $I \in R^{H \times W \times 3}$ and textual driving context $T$ (navigation instructions, traffic rules). The output consists of structured driving decisions $y$ representing waypoints, actions, and control commands. Mathematically, our model learns the conditional distribution $P_\theta(y|I, T)$ through the diffusion framework.

\subsection{Vision Encoder and MLP Connector}
The vision encoder processes camera inputs to extract spatial-temporal features, which are then projected into the language model's embedding space for multimodal processing. The SigLP-2~\cite{tschannen2025siglip2multilingualvisionlanguage} is used as the base vision encoder to process input images $I$, producing visual tokens $V \in R^{N_v \times d_{model}}$, where $N_v$ represents the number of visual tokens and $d_{model}$ is the hidden dimension. Then, the visual features $V$ are projected into the language model's embedding space through a two-layer MLP connector, and the projected visual tokens $V_{project} = MLP(V)$ are concatenated with textual driving context tokens $T_{emb}$ to form a unified multimodal input sequence:

\begin{equation}
    X_{input} = \text{Concat}([V_{project}, T_{emb}]),
\end{equation}
where $T_{emb}$ represents the embedded textual driving context.

\subsection{Diffusion Language Backbone}
The core of ViLaD is a masked diffusion transformer that processes the concatenated multimodal tokens through bidirectional self-attention, enabling joint reasoning over visual and textual information.

\subsubsection{Fix Pattern Training}

The backbone employs a bidirectional transformer without causal masking, processing the concatenated multimodal sequence through transformer layers. The bidirectional self-attention enables the model to consider the complete context from both visual and textual modalities when predicting individual driving decisions.

During training, we utilize the masked diffusion approach specifically adapted for multimodal conditional generation and end-to-end autonomous driving tasks. 
Specifically, our input is a multimodal input $X_{input}$ consisting of visual features $V$ and textual driving context $T$.
The target output $y_0$ is a fixed-pattern driving sequence that includes both control actions and punctuation marks, where punctuation marks always appear in predefined positions and the length of the $y_0$ is always consistent. This design ensures the model focuses solely on predicting action values rather than unnecessary formatting. We construct the model input as $X_t = \text{Concat}([X_{\text{input}}, y_0])$, applying masks only to the action tokens in $y_0$ while keeping $X_{\text{input}}$ and the punctuation tokens visible throughout training.


In each training step, for a training sample $X_{t} = \text{Concat}([X_{input}, y_0])$ , we sample a noise level $t  \sim U(0,1)$ and independently mask each token in the target action tokens with probability t:
\begin{equation}
    q(y_t|y_0) = \prod_{i=1}^{L} q(y_t^i|y_0^i),
\end{equation}
where $L$ is the length of the driving sequence $y_0$. The projected visual features $V_{project}$, textual context $T$ and punctuation marks remain unmasked throughout the process.

The training objective is:

\begin{equation}
    \mathcal{L}(\theta) = -\frac{1}{t} \sum_{i=1}^{L} \mathbb{1}[y_t^i = \text{[M]}] \log p_{\theta}(y_0^i | X_t),
\end{equation}
where $\mathbb{1}[y_t^i = \text{[M]}]$ ensures the loss is computed only on masked driving decision tokens, and $X_t$ represents the full concatenated sequence with masked driving responses.


\algrenewcommand\algorithmicrequire{\textbf{Input:}}
\algrenewcommand\algorithmicensure{\textbf{Output:}}

\begin{algorithm}[htbp]
\small
\caption{Remasking Process}\label{alg:remask}
\begin{algorithmic}[1]

    \Require Current masked sequence $y_t$, predictions $\hat{y}_0$, total number of tokens $L$, total steps $S$, threshold $\tau$
    \Ensure Next sequence state $y_s$

    
    \State $\textit{confidence}_{i} \gets p_{\theta}(\hat{y}_{0}^{i} \mid X_{t})$ for $i = 1, \dots, L$
    \State $\text{masked\_indices} \gets \{i \mid y_{t_i} = \text{[M]}\}$
   
    \State $n_{\text{unmask\_schedule}} \gets \lceil \frac{L}{S} \rceil$
    \Comment{Number of tokens to unmask based on schedule}
    
    \State $n_{\text{unmask\_threshold}} \gets |\{i \in \text{masked\_indices} \mid \textit{confidence}_i > \tau \}|$
    \Comment{Number of tokens to unmask based on confidence}
    
    \State $n_{\text{unmask}} \gets \max(n_{\text{unmask\_schedule}}, n_{\text{unmask\_threshold}})$
    
    \State $\text{conf}_{\text{masked}} \gets \{i \in \text{masked\_indices}| \textit{confidence}_i \in \text{topk}(\textit{confidence}, n_\text{unmask})\}$ \Comment{Get top-n confident tokens}
    \For{$i \in \text{conf}_{\text{masked}}$}
        \State $y_{s}^{i} \gets \hat{y}_{0}^{i}$ \Comment{Unmask the selected predictions}
    \EndFor
\end{algorithmic}
\end{algorithm}

\begin{table*}[!t]
    \centering
    \small
    \caption{End-to-end planning performance on the nuScenes dataset (* results from ~\cite{xing2025openemma}).}
    \resizebox{\linewidth}{!}{
    \begin{tabular}{l|c|llllll}
        \toprule
        Method & Learning & L2 (m) 1 s ($\downarrow$) & L2 (m) 2 s ($\downarrow$) & L2 (m) 3 s ($\downarrow$) & L2 (m) Avg ($\downarrow$) & Failure Rate (\%) ($\downarrow$)\\
        \midrule
        LLaVA-1.6-7B* & \multirow{4}{3em}{Zero-Shot} 
        & 1.66 & 3.54 & 4.54 & 3.24 & 4.06 \\
        Llama-3.2-11B* & 
        & 1.50  & 3.44 & 4.04 & 3.00  & 23.92 \\
        Qwen2-VL-7B* & 
        & 1.22 & 2.94 & 3.21 & 2.46  & 24.00\\
        \rowcolor{black!10} ViLaD-Zero & 
        & 1.09 & 2.51 & 3.74 & 2.45 & 0.03  \\
        \hline
        LLaVA-1.6-7B* & \multirow{4}{3em}{Open-EMMA} 
        & 1.49 & 3.38 & 4.09 & 2.98 & 6.12  \\
        Llama-3.2-11B* & 
        & 1.54 &3.31&3.91 & 2.92 & 22.00  \\
        Qwen2-VL-7B* & 
        & 1.45 &3.21 &3.76 &2.81 & 16.11 \\
        \rowcolor{black!10} ViLaD-CoT & 
        & 1.24 & 2.74 & 4.13 & 2.71  & 0.17 \\
        \hline
        LLaVA-1.6-7B & \multirow{4}{3em}{SFT} 
        & 0.91 & 2.50 & 3.44 & 2.28 & 55.25 \\
        Llama-3.2-11B & 
        & 0.80 & 2.31 & 3.10 & 2.07 &0.06  \\
        Qwen2-VL-7B & 
        & 1.32 & 2.94 & 3.98 & 2.74 & 0.03 \\
        \rowcolor{black!10} ViLaD-SFT & 
        & \textbf{0.79} & \textbf{1.92} & 2.83 & 1.85 & \textbf{0.00}  \\
        \hline
        \rowcolor{black!10} ViLaD-Opt & Efficient Optimized SFT
        & 0.81 & 1.93 & \textbf{2.69} & \textbf{1.81}  & \textbf{0.00} \\
        \bottomrule
    \end{tabular}
    }
    \label{tab:main-result}
    \vspace{-6mm}
\end{table*}

\subsubsection{Fix Pattern Confidence-based Inference}

During inference, we perform the reverse diffusion process starting from a partially masked sequence $y_1$, where all target action tokens are masked while punctuation tokens and length remain fixed, consistent with the training setup. The model then iteratively refines the masked tokens over up to $S$ denoising steps to generate the final driving action sequence.

One primary contribution we achieve is focusing on addressing quality degradation in parallel token generation caused by the conditional independence assumption in diffusion LLMs. When unmasking multiple token positions $i$ and $j$, MDMs sample these from $p_\theta(y^i_0|y_t) \cdot p_\theta(y^j_0|y_t)$ due to the conditional independence assumption, while the true joint probability is $p_\theta(y^i_0, y^j_0|y_t) = p_\theta(y^i_0|y_t) \cdot p_\theta(y^j_0|y_t, y^i_0)$. This discrepancy degrades generation quality. To address this, enlightened by Fast-dLLM~\cite{wu2025fastdllmtrainingfreeaccelerationdiffusion}, we propose confidence-aware parallel decoding: at each iteration, we compute confidence scores for each token and only unmask those exceeding the threshold $\tau$. If the number of tokens with confidence scores exceeding $\tau$ is less than $\lceil L/S \rceil $, we unmask the $\lceil L/S \rceil $ highest-confidence tokens to ensure progress.

The inference steps begin with sampling. At a step with the current sequence $X_t$, we use the model to predict all masked tokens simultaneously:

\begin{equation}
    \hat{y}_{0}^i = \underset{y_{0}^i}{\arg\max} \, p_{\theta}(y_{0}^i \mid X_{t}) \quad \text{for all } i.
\end{equation}

Then, following the threshold-based remasking strategy, we keep at least $\lceil L/S \rceil$ high-confidence predictions unchanged, and remask all the other tokens. The remask process is summarized in Algorithm \ref{alg:remask}.
After remasking, the resulting sequence is used as the input sequence in the next step. 
This iterative process lasts for at most $S$ steps until all tokens are masked, following the standard discretized reverse diffusion schedule. Both the number of sampling steps $S$ and the threshold $\tau$ can be adjusted to balance between generation quality and inference speed. The whole inference process is summarized in Algorithm \ref{alg:vilad_compact_modern}.

\begin{algorithm}[htbp]
\small
\caption{ViLaD Inference}\label{alg:vilad_compact_modern}
\begin{algorithmic}[1]
    \Require Visual features $V$, textural driving context $T$, driving sequence length $L$, steps $S$
    \Ensure Predicted driving sequence $y_0$
    \State Initialize: $X_1 \leftarrow \text{Concat}([\text{MLP}(\text{VE}(V)), \text{TE}(T), [\text{[M]}]^L])$
    \While{$y_s$ contains [M]} \Comment{Loop until all tokens are unmasked}
        \State $\hat{y}_0 \leftarrow \arg\max_{y_0} p_\theta(y_0 | X_t)$ \Comment{Parallel prediction}
        \State $y_s \leftarrow \text{Remask}(y_t, \hat{y}_0, p_\theta(\hat{y}_0^i | X_t))$ \Comment{Remask Process}
        \State $X_t \leftarrow \text{Concat}([V_{project}, T_{emb}, y_s])$
    \EndWhile
    \State \Return $y_0$
\end{algorithmic}
\end{algorithm}

\section{Experiment and Results}
\subsection{Experiment Setup}

We conduct experiments on the nuScenes dataset~\cite{caesar_nuscenes_2020} using only front camera images from the last frame and ground-truth trajectories for training and validation. Evaluation focuses on L2 error in the first 3 seconds and inference latency for real-time performance. Followed Open-EMMA~\cite{xing2025openemma}, a predic-
tion is considered a failure if the L2 norm exceeds 10 within the first second of the future trajectory. We compare ViLaD against three state-of-the-art VLMs LLaVA-1.6-Mistral-7B~\cite{li_llava-med_2023}, Llama-3.2-11B-Vision-Instruct~\cite{zhang_video-llama_2023}, and Qwen2-VL-7B-Instruct~\cite{wang2024qwen2vlenhancingvisionlanguagemodels} zero-shot, OpenEMMA prompting~\cite{xing2025openemma}, and supervised fine-tuning (SFT) settings. 

 We evaluate multiple configurations of our proposed ViLaD framework to analyze different aspects of performance and capabilities. LLaDA-V~\cite{you2025lladavlargelanguagediffusion} is our base model. Other ViLaD variants include ViLaD-SFT (full model with diffusion-based SFT), ViLaD-Zero (zero-shot), ViLaD-CoT (chain-of-thought reasoning), and ViLaD-Opt (optimized inference), offering a broad evaluation of performance and capabilities.

\subsection{Optimization Strategy}

The outcome from dLLM-Cache~\cite{liu2025dllmcacheacceleratingdiffusionlarge}dLLM-Cache~\cite{liu2025dllmcacheacceleratingdiffusionlarge} introduces a training-free adaptive caching framework that combines long-interval prompt caching with partial response updates guided by feature similarity. This methods accelerate the inference speed of ViLaD and hence reduce the latency of the whole framework.

\subsection{End-to-End Planning}
Table~\ref{tab:main-result} presents comprehensive comparison results across different methods. Our ViLaD framework demonstrates the best performance across all evaluation metrics compared to baseline VLM approaches. In the zero-shot setting, ViLaD-Zero achieves remarkable performance with L2 errors of 1.09 m, 2.51 m, and 3.74 m at 1 s, 2 s, and 3 s horizons, respectively, significantly outperforming the best baseline (Qwen2-VL-7B~\cite{wang2024qwen2vlenhancingvisionlanguagemodels}), which achieves 1.22 m, 2.94 m, and 3.21 m. Most notably, ViLaD-Zero maintains an extremely low failure rate of 0.03\% compared to baseline failure rates ranging from 4.06\% to 24.00\%, demonstrating the inherent robustness of our diffusion-based approach.

The OpenEMMA prompting framework shows consistent advantages of our approach. ViLaD-CoT achieves competitive performance with an average L2 error of 2.71 m while maintaining a near-zero failure rate (0.17\%), outperforming all baseline methods in this category. However, we found that the ViLaD model shows no improvement using the chain-of-thought prompting method. The chain-of-thought prompting method still needed to be explored.

With only the SFT setting (no fixed pattern training and fixed pattern confidence threshold inference), ViLaD-SFT achieves the best overall performance with L2 errors of 0.79 m, 1.92 m, and 2.83 m across the three time horizons, resulting in an average L2 error of 1.85 m with zero failure rate. This represents significant improvements over the best baseline (Llama-3.2-11B~\cite{zhang_video-llama_2023} SFT), which achieves 2.07 m average L2 error. Notably, while some baseline methods achieve competitive L2 errors, they suffer from substantially higher failure rates (up to 55.25\% for LLaVA-1.6-7B~\cite{li_llava-med_2023} SFT), highlighting the safety advantages of our approach.

ViLaD-Opt, our optimized deployment configuration with fixed pattern training and fixed pattern confidence threshold inference, maintains the most competitive performance (1.81 m average L2 error) while achieving inference speedup, making it suitable for real-time autonomous driving applications.

\begin{figure}[!t]
    \centering
        \includegraphics[width=0.9\linewidth]{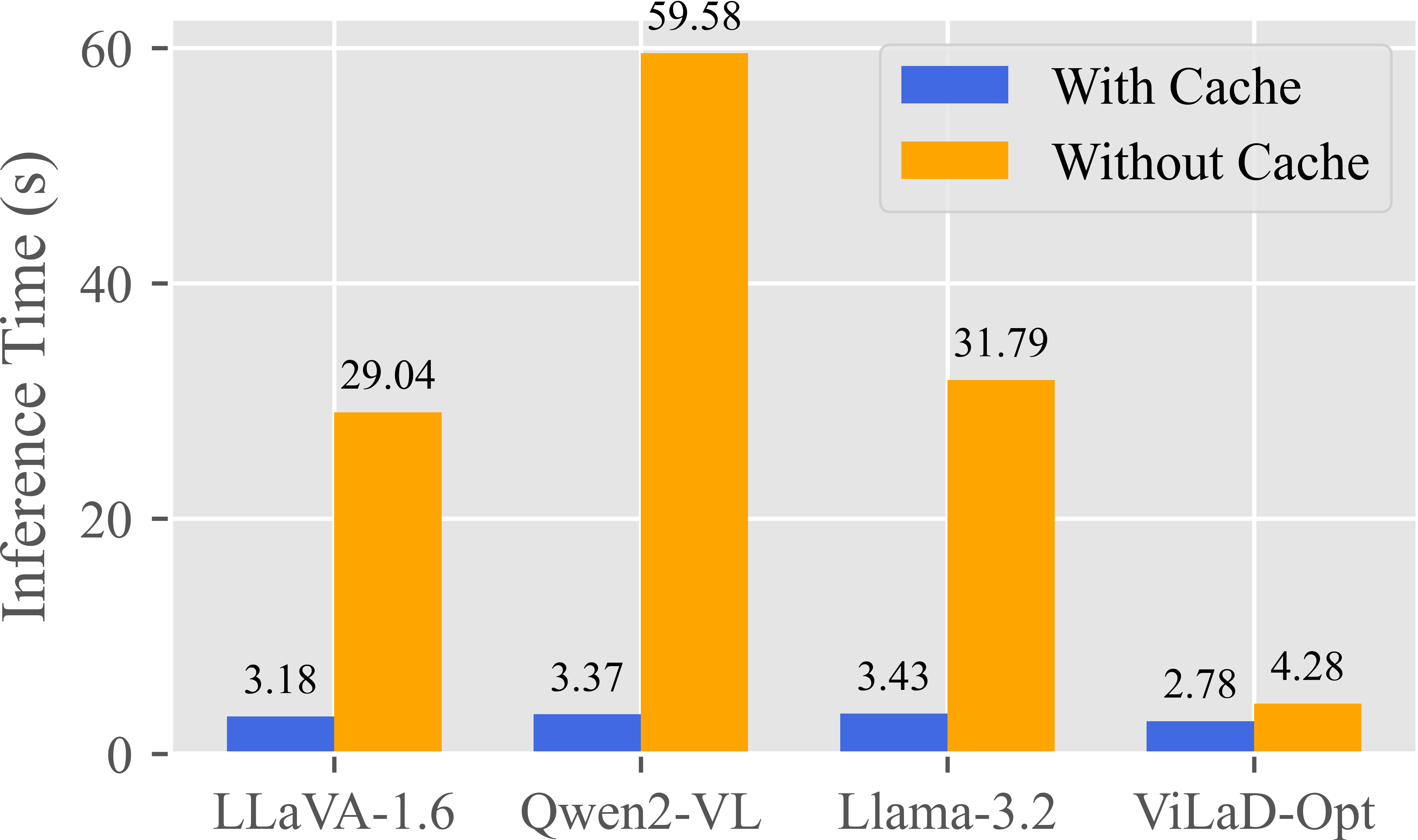}
    \vspace{-2mm}
    \caption{The comparison of inference time between VLMs and ViLaD-Opt.}
    \label{fig:te}
    \vspace{-6mm}
\end{figure}

\subsection{Inference Efficiency Experiment}
To validate the real-time performance of our framework, we conducted a comprehensive inference efficiency experiment, pitting our approach against leading autoregressive VLM baselines that benefit from highly mature optimization technologies. The VLM baselines (LLaVA-1.6-7B~\cite{li_llava-med_2023}, Qwen2-VL-7B~\cite{wang2024qwen2vlenhancingvisionlanguagemodels}, and Llama-3.2-11B~\cite{zhang2023videollamainstructiontunedaudiovisuallanguage}) were accelerated using standard KV caching, while our ViLaD-Opt model, representing a still-developing class of vision language diffusion models, utilized a dLLM cache. The results from the experiment, conducted on a single NVIDIA A100 GPU and presented in Figure \ref{fig:te}, reveal the significant computational advantages of our diffusion-based method. Without caching optimizations, ViLaD-Opt achieves an inference time of 4.28 s, proving to be 6.5 to 14 times faster than the autoregressive models, which required between 28.10 s and 60.18 s. While caching accelerated all models, ViLaD-Opt maintained its superiority with an inference time of 2.78 s, still outperforming the fastest baseline (3.18 s). This experiment empirically confirms that the parallel generation paradigm of ViLaD provides a critical advantage in inference speed by avoiding the sequential, token-by-token processing bottleneck inherent in autoregressive architectures, making it highly suitable for deployment in safety-critical, real-time applications like autonomous driving.

\begin{figure}[!t]
    \centering
        \includegraphics[width=\linewidth]{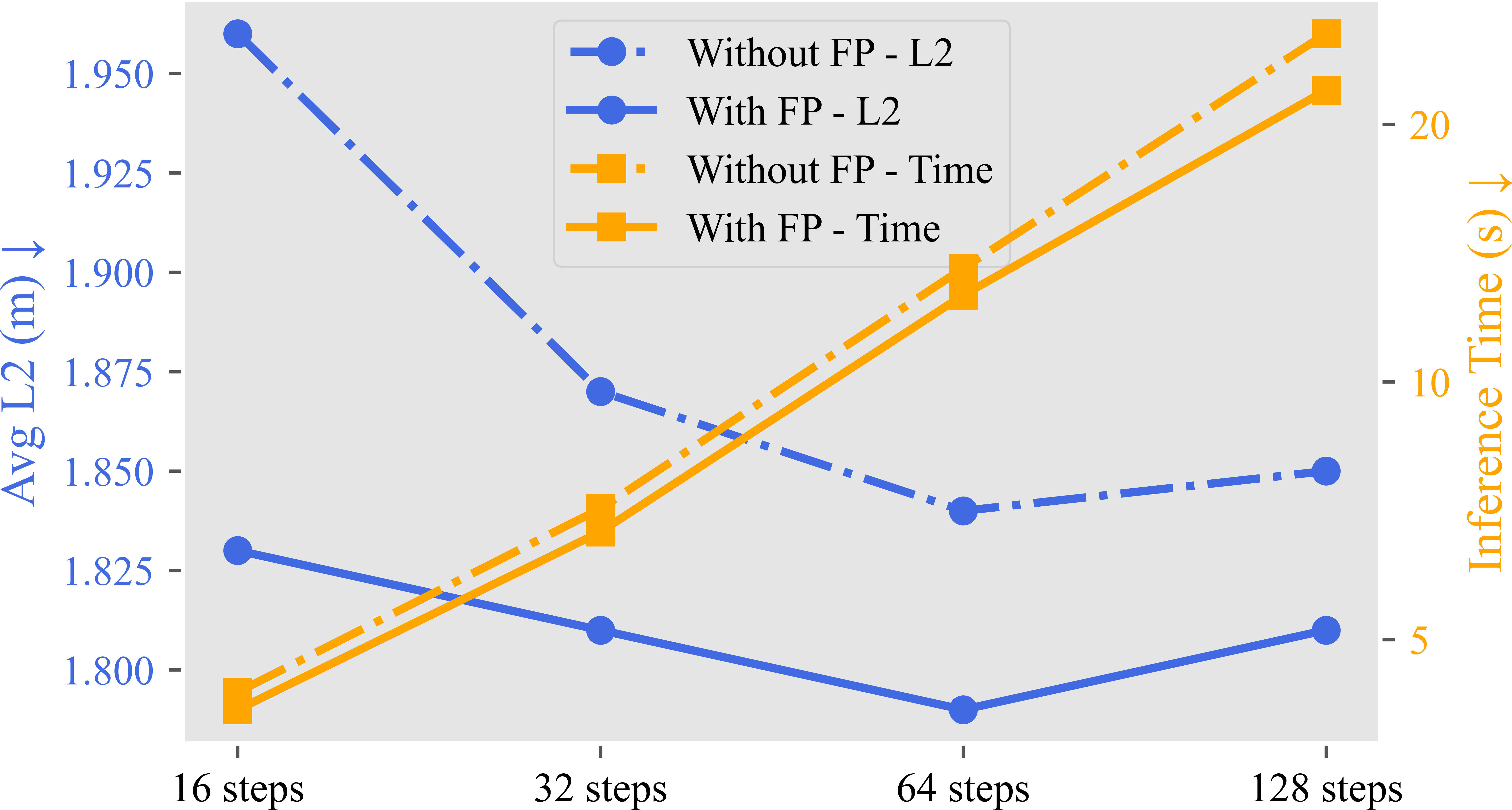}
    \vspace{-6mm}
    \caption{The effectiveness of our Fixed Pattern (FP) optimization strategy across different numbers of diffusion steps.}
    \label{fig:ab_FB}
    \vspace{-6mm}
\end{figure}

\begin{table}[b!]
\caption{The ablation study of our confidence threshold optimization strategy across different thresholds.}
\centering
\resizebox{0.9\linewidth}{!}{%
\begin{tabular}{c|cc}
\toprule
Confidence Threshold & Avg L2 (m) $(\downarrow)$ & Inference Time (s) $(\downarrow)$\\
\midrule
0.9 & 1.82  & 3.95\\
0.7 & 1.87 & 3.29\\
0.5 & 1.81 & 2.78\\
0.3 & 1.97 & 2.73\\
\bottomrule
\end{tabular}
}
\label{tab:abt}
\vspace{-6mm}
\end{table}


\begin{table*}[t]
\caption{On-Board Model Performance Metrics for Interactive Parking.}
\centering
\resizebox{0.9\textwidth}{!}{%
\begin{tabular}{c|cc|cc|cc}
\toprule
\multirow{2}{*}{Metrics} & \multicolumn{2}{c|}{Model Performance} & \multicolumn{2}{c|}{Safety Metrics} & \multicolumn{2}{c}{Comfort Metrics} \\
\cline{2-7}

& \begin{tabular}[c]{@{}c@{}}Latency (s)\end{tabular} & \begin{tabular}[c]{@{}c@{}}Success Rate (\%)\end{tabular} & \begin{tabular}[c]{@{}c@{}}Long. Vel. Var.($10^{-4}$ m²/s²)\end{tabular} & \begin{tabular}[c]{@{}c@{}}Lat. Vel. Var.($10^{-4}$ m²/s²)\end{tabular} & \begin{tabular}[c]{@{}c@{}}Max Long. Accel.(m/s²)\end{tabular} & \begin{tabular}[c]{@{}c@{}}Max Lat. Accel.(m/s²)\end{tabular} \\
\midrule
\textit{Results} & 0.96 & 93.51 & 1.07 & 2.66 & 1.43 & 1.12 \\
\bottomrule
\end{tabular}%
}
\label{tab:onboard_performance}
\vspace{-6mm}
\end{table*}

\subsection{The Effectiveness of the Fix Pattern}


As shown in Figure~\ref{fig:ab_FB}, without Fixed Pattern optimization, increasing diffusion steps from 16 to 128 improves L2 error from 1.96 m to 1.85 m but increases inference time from 4.34 s to 25.52 s. This trade-off makes higher step counts impractical for real-time applications. 
With FP optimization, we achieve huge speedup and improved accuracy across all step configurations. For example, at 16 steps, inference time reduces from 4.34 s to 3.12 s (28\% improvement) while we achieve better L2 error (1.85 m vs 1.96 m). At 128 steps, the optimization reduces inference time from 25.52 s to 21.23 s (17\% improvement) while also improving accuracy (1.78 m vs 1.85 m).


The results indicate that Fixed Pattern optimization provides consistent benefits, with the most improvements at lower step counts. For practical deployment, 32 steps with Fixed Pattern optimization provide an optimal balance between accuracy and efficiency, achieving under 2 m L2 error with inference times under 3 seconds.

\subsection{Ablation Study on Confidence Threshold}

We conduct detailed ablation studies to understand the impact of key hyperparameters on system performance. The confidence threshold analysis shows important insights about the trade-off between accuracy and efficiency in our confidence-based remasking strategy. As shown in Table~\ref{tab:abt}, a threshold of 0.5 achieves the best balance with 1.81 m average L2 error and 2.78 s inference time. Higher thresholds (0.9, 0.7) tend to be more conservative, unmasking fewer tokens per iteration, which can lead to slightly higher accuracy (0.9: 1.82 m, 0.7: 1.87 m) but at the cost of increased inference time (0.9: 3.95 s, 0.7: 3.29 s). Lower thresholds (0.3) unmask more tokens aggressively, reducing inference time to 2.73 s but potentially degrading accuracy to 1.97 m due to premature commitment to low-confidence predictions.


\section{Real-World Case Studies: On-Board Interactive Parking}
To further validate the real-world applicability of our approach, we conduct a case study focused on deploying a smaller, specialized version of our model for an interactive parking task. The objective is to demonstrate the model's ability to perform real-time, on-board decision-making based on natural language commands within a constrained vehicle hardware environment.

\subsection{Vehicle and System Setup}
\begin{figure}[!t]
    \centering
    \includegraphics[width=\linewidth]{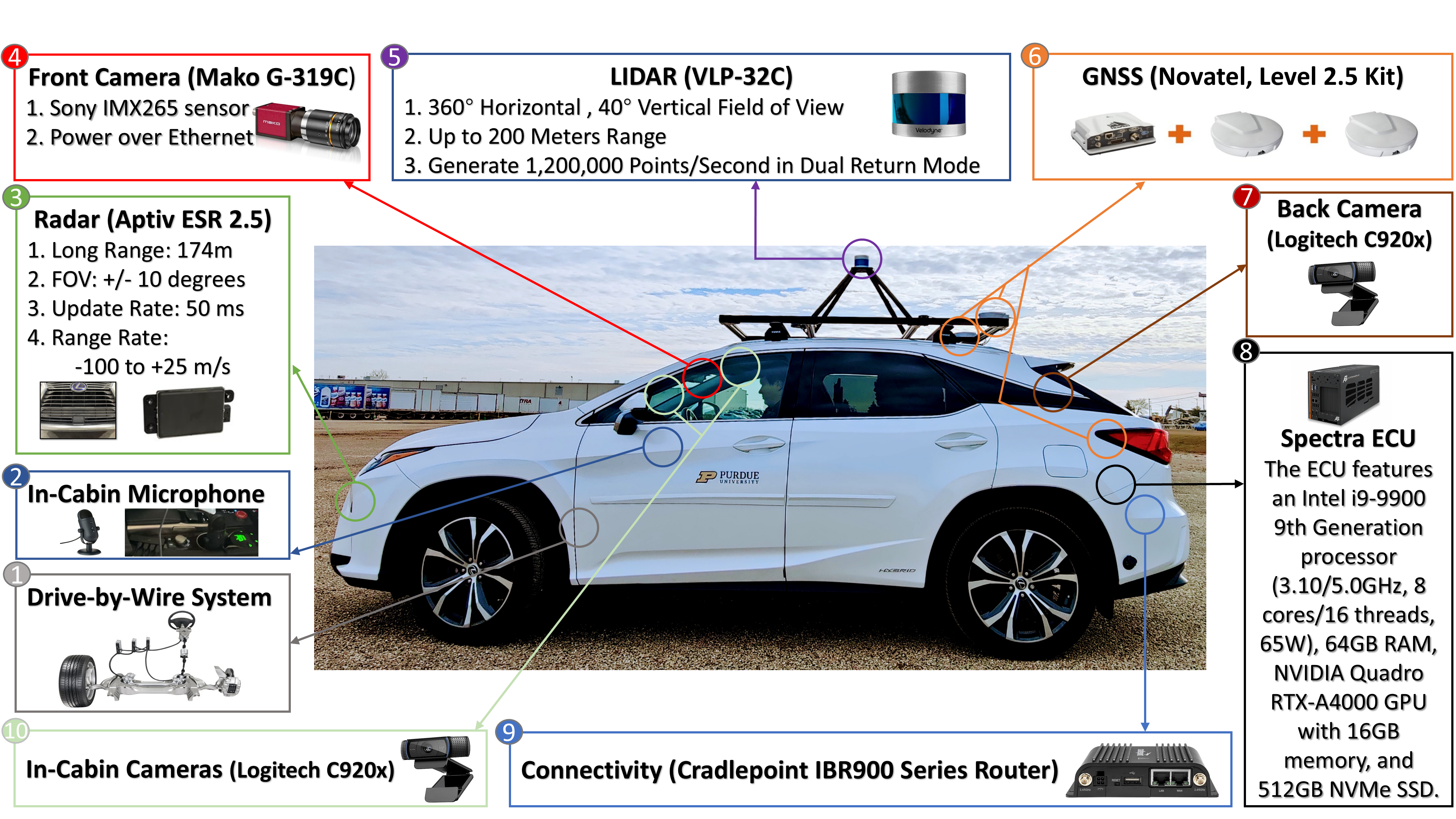}
    \caption{Overview of the autonomous vehicle setup.}
    \label{fig:vehicle_setup}
    \vspace{-6mm}
\end{figure}

The experiment is performed on a production vehicle equipped with a custom sensor suite and computational hardware, as illustrated in the vehicle setup Figure~\ref{fig:vehicle_setup}. The on-board processing is handled by a Spectra ECU featuring an Intel i9-9900 CPU and an NVIDIA Quadro RTX-A4000 GPU. Inputs for the task are captured by in-cabin cameras with microphones, which receive the driver's commands. The system leverages a pre-recorded map of the parking area and integrates with the vehicle's existing autonomous driving stack for final motion planning and control.

\subsection{Task and Methodology}
The task requires the model to select a preferred parking spot in a lot based on a human's spoken command (e.g., ``Find a spot away from other cars''). The model's direct output is the final waypoint corresponding to the chosen parking spot, which is then fed to the vehicle's planner. This setup allows us to isolate and evaluate the model's language understanding and decision-making performance.

We train a specialized model based on the SMDM~\cite{nie2025scalingmaskeddiffusionmodels} architecture for 6 epochs using a self-created dataset of 3033 command-action pairs. We adopt SMDM as the foundation for our real-world case study due to its lightweight and deployable architecture, while still using key techniques such as classifier-free guidance, which are later scaled up and adopted by LLaDA, the backbone of our ViLaD framework. This dataset is meticulously created, where the first 195 pairs are created manually, and the remaining 2,838 are generated by GPT-4o~\cite{openai_gpt-4_2023}. To ensure high quality, each generated pair is then judged for correctness by another LLM (Gemini~\cite{team2023gemini}) and a human checker. If the AI and human annotator can not reach an agreement, we recruit more people to vote on the correct pair, guaranteeing the quality of our data.

\subsection{Results and Discussion}

The on-board model demonstrates excellent performance across all evaluation metrics, as shown in Table~\ref{tab:onboard_performance}. The system achieves an average inference latency of just 0.96 s, confirming its capability for real-time human-vehicle interaction. The task is completed with a 95.67\% success rate, indicating high reliability in correctly interpreting driver intent and selecting a valid parking spot.


Furthermore, the vehicle's dynamic performance during the maneuvers highlights the quality of the model's decisions. Safety metrics are strong, with a low longitudinal velocity variance of 1.07 and a lateral velocity variance of 2.66 $\times 10^{-4}$ $m^2/s^2$, making stable and predictable vehicle control. Comfort metrics also remain high, with a maximum longitudinal acceleration of \SI{1.43}{\meter\per\second\squared} and a maximum lateral acceleration of \SI{1.12}{\meter\per\second\squared}, ensuring a smooth experience for the occupants.

This case study successfully demonstrates that a specialized vision-language diffusion model can be effectively deployed on an embedded vehicle platform. The model not only performs its primary decision-making task with high accuracy and low latency but also contributes to a safe and comfortable autonomous driving experience.

\section{Conclusion}



We present ViLaD, a novel end-to-end autonomous driving framework that replaces conventional autoregressive models with a Large Vision Language Diffusion (LVLD) approach. By leveraging masked diffusion, ViLaD enables parallel action generation, includes bidirectional context, and adopts an easy-first decision strategy to reduce latency and improve robustness. Experiments on the nuScenes benchmark demonstrate that ViLaD outperforms state-of-the-art vision-language models in both accuracy (1.82 m average 2 error) and efficiency (3-second planning time), achieving near-zero failure rates in dynamic scenarios. The successful on-board deployment for a real-world interactive parking task further proves the practical value (93.51\% success rate) and real-time performance (0.96 s) of our approach, underscoring its potential for real-world automotive applications.


\bibliography{bib/Int_LLM,bib/jiaru,bib/can,bib/zichong,bib/JP,bib/previous,bib/previous2}
\bibliographystyle{IEEEtran}

\end{document}